\documentclass{cta-author}

\usepackage{booktabs} 
\usepackage{graphicx}
\usepackage{epstopdf}
\usepackage{subfigure}
\usepackage{amsmath,amssymb,amsfonts}
\usepackage{xcolor}
\usepackage{enumerate}
\usepackage{amsthm}
\usepackage{diagbox}
\usepackage{enumitem} 
\usepackage{color,colordvi}
\usepackage{textcomp}
\usepackage{epsfig}
\usepackage{url}
\usepackage{mdwlist}
\usepackage{makecell}
\usepackage{verbatim}
\usepackage{multirow,multicol}
\usepackage{setspace}
\usepackage{float}
\usepackage{fancyhdr}
\usepackage[normalem]{ulem}
\usepackage{makecell}
\usepackage{fancyhdr}
\usepackage{flushend}
\usepackage{algorithm}
\usepackage{algorithmic}
\usepackage{natbib}
\usepackage{multirow}
\usepackage{soul}
\usepackage{stfloats}
\usepackage[colorlinks,linkcolor=green,citecolor=green]{hyperref}
{}
{}
{}

\begin{document}
\title{Enhanced Decentralized Federated Learning based on Consensus in Connected Vehicles}

\author{\au{Xiaoyan Liu$^{1}$}, \au{Zehui Dong$^{1}$}, \au{Zhiwei Xu$^{1,2\corr}$}, \au{Siyuan Liu$^{1}$}, \au{Jie Tian$^{3}$},   
}

\address{\add{1}{College of Data Science and Application, Inner Mongolia University of Technology, Huhhot, China, 010080}
\add{2}{Institute of Computing Technology, Chinese Academy of Sciences, Beijing, China, 100190}
\add{3}{Department of Computer Science, New Jersey Institute of Technology, NJ, USA, 07102}
\email{xuzhiwei2001@ict.ac.cn}}

\begin{abstract}
Advanced researches on connected vehicles have recently targeted to the integration of vehicle-to-everything (V2X) networks with Machine Learning (ML) tools and distributed decision making. Federated learning (FL) is emerging as a new paradigm to train machine learning (ML) models in distributed systems, including vehicles in V2X networks. Rather than sharing and uploading the training data to the server, the updating of model parameters (e.g., neural networks’ weights and biases) is applied by large populations of interconnected vehicles, acting as local learners. Despite these benefits, the limitation of existing approaches is the centralized optimization which relies on a server for aggregation and fusion of local parameters, leading to the drawback of a single point of failure and scaling issues for increasing V2X network size. Meanwhile, in intelligent transport scenarios, data collected from onboard sensors are redundant, which degrades the performance of aggregation. To tackle these problems, we explore a novel idea of decentralized data processing and introduce a federated learning framework for in-network vehicles, C-DFL( Consensus based  Decentralized Federated Learning), to tackle federated learning on connected vehicles and improve learning quality. Extensive simulations have been implemented to evaluate the performance of C-DFL, that demonstrates C-DFL outperforms the performance of conventional methods in all cases.
\end{abstract}

\maketitle

\section{INTRODUCTION}\label{sec1}

\begin{figure*}[h!]
  \begin{center}
  \includegraphics[width=0.95\textwidth]{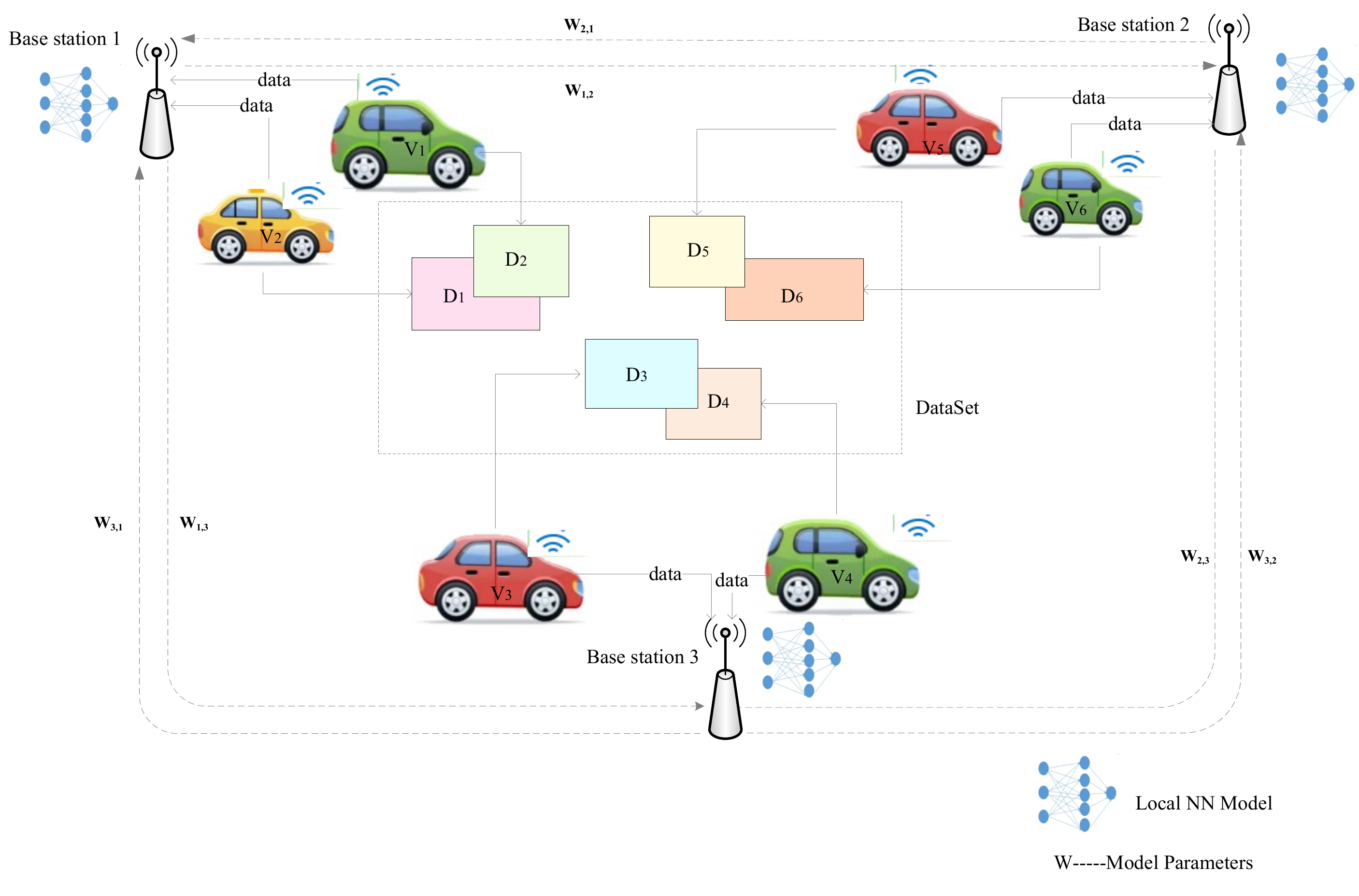}
  \caption{Decentralized FL architecture: vehicles transmit the data to base stations, which then share parameters to collaboratively train the ML model.
  }\label{fig1}
  \end{center}
\end{figure*}
A new technical analysis by the National Highway Traffic Safety Administration (NHTSA) states that human error is to blame for 94$\%$ percent of traffic accidents\cite{1}. Autonomous driving vehicles (AVs) with V2X links can also free labor from simple, repetitive driving by relying on artificial intelligence, visual computing, radar, surveillance devices, and global positioning systems to work together to allow computers to operate motor vehicles autonomously, which can prevent accidents caused by human driving errors. The AVs are expected to perceive the surroundings with data captured by a variety of onboard sensors in near-real-time \cite{2}. However, the increasing data collected by AVs become an issue holding back the development of self-driving technology. The emergence of distributed deep learning (DML) will alleviate this situation. 

Distributed Machine Learning is currently one of the most popular research fields in machine learning. Due to the good flexibility and scalability of DML, single-machine resources can be effectively combined. Hence, connected automated driving (CAD) ushered in a new development. Cooperative multi-vehicle control and planning strategies are the emphasis of the CAD functions \cite{3}. Nevertheless, distributed training on autonomous driving vehicles is limited by the following problems. The first is the huge burden that the Internet brings to the backbone network to transmit raw data. Furthermore, it is impossible to share such vast amounts of data. 

Federated Learning (FL), which offers improved privacy-preserving functions in comparison to DML systems \cite{4}, has been developing in recent years to handle large-scale distributed training across numerous linked devices or agents.
Furthermore, FL has recently started to gain attention in connected automated driving applications. With FL, autonomous vehicles send model updates to a so-called parameter server rather than sharing raw sensor data with the server training the model \cite{5}. FL avoids the need for the data to leave the edge devices(refers to the base station in this paper), improving privacy, lowering the computational burden for the server, and decreasing the communication overhead when the model update is smaller than the data to be transmitted per iteration. In early FL implementations, research of FL focuses on Centralized Federated Learning (CFL). However, this approach comes with significant drawbacks \cite{6,7,8} and scaling issues for increasing V2X network size. Not only does a parameter server create a single point of failure vulnerable to crashes or hacks, but it can also become a performance bottleneck as the number of devices pushing model updates increases, which makes it difficult for autonomous driving to process data in real-time.

This drives researches on decentralized federated learning (DFL) in AVs (as depicted in Fig.1). Decentralized solutions to FL based on a distributed implementation of SGD \cite{9} have been thus proposed. As shown in the example of Fig.1, base station $1$ receives the model updates parameters from the neighbors base station $2$ and base station $3$. Then, it upgrades the local parameters. Models are trained using a decentralized topology and the parameter server is removed with DFL. With DFL, base stations rely on local cooperation with neighbors, each base station is connected to a subset of the other base stations in the network from which it receives incoming models and to which it pushes its updated models. DML solves the communication bottleneck to a certain extent. Decentralized solutions are therefore favored in intelligent transport scenarios, and local processing makes it possible to accelerate the learning process. Additionally, the AVs perceive the surroundings via analyzing a large amount of data captured by a variety of onboard sensors in near-real-time. More especially, data collected by onboard sensors may be redundant, which affects the performance of aggregation.

To solve the aforementioned problem, we should constitute a new aggregation strategy to train a model with data from sensors. In this work, we propose  C-DFL, a novel idea of decentralized data processing and federated learning framework for in-network vehicles, which considers redundant data. The main contributions are summarized as follows:

(1)	To satisfy new intelligent transport scenarios, we propose a decentralized federated learning framework, C-DFL, for distributed environment understanding of in-network vehicles.

(2) Considering the redundant data collected from vehicles, we explore a novel aggregation paradigm of local model updates, by mapping data into a compact representation as a record of local data distribution. Through exchanging this type of compact representations, base stations can filter redundant data and thus speed up local updating and update aggregation. 

(3)	Using two real-world datasets, we implement our method and baselines on an NS-3-based simulation platform. Extensive simulations demonstrate that our method outperforms all baselines in terms of accuracy as well as convergence speed. 

This paper is organized as follows: We review the related work about DFL in Section 2, and detail the system model of decentralized FL training in Section 3.
In Section 4, a decentralized federated learning framework is proposed with a novel aggregation paradigm of local model updates to reduce the impact of redundant data to different local models.
Two real-world datasets is used to demonstrate the proposed framework outperforms the state of the art in a V2X scenario. 
Finally, we conclude this work in Section 6.

\section{RELATED  WORK}\label{sec2}

New types of decentralization are anticipated to support next-generation networks. Devices can work together directly over device-to-device (D2D) spontaneous connections thanks to these networks, which are created without the assistance of a central coordinator. D2D techniques bring further advances for Decentralized FL. Instead of depending on centralized solutions, collaborating devices in Decentralized FL share model parameters via D2D connections and set a consensus policy into place. Edge nodes don't need to rely on a central server for fast training parameters feedback, reducing the communication bottleneck of the FL.

Devices sample, convergence, and stochastic heterogeneity are the three basic difficulties in DFL design \cite{10,11,12,13}. Reduced FL process convergence times are crucial, especially in mobility scenarios involving autonomous vehicles.

To reduce the communication bottleneck of the FL, some studies focus on decentralized FL. Lan et al. \cite{14} provided a decentralized stochastic algorithm. Sirb and Ye \cite{15} gave an asynchronous decentralized stochastic algorithm. However, those algorithms are provided not accelerated. For the purpose of improving the communication efficiency of FL, Gossip-based protocol for distributed learning has been explored in the data center setting as an alternative to the parameter-server approach \cite{16}. However, when the communication speed of the nodes and the heterogeneity of the data are related, the GL cannot converge. Later, Guha et al. \cite{17} addressed a decentralized FL based on segmented gossip. Lu et al. \cite{18} applied the decentralized FL to electronic health records. 
These studies demonstrate how, under certain circumstances, transitioning from a centralized sharing scheme to a decentralized one might enable models to establish a consensus at the global minimum while avoiding server node-related communication delays. In  \cite{19}, a segmented gossip aggregation is proposed. However, it’s extremely application dependent and not suitable for more general ML contexts. More recently, Savazzi et al.\cite{20} proposed a consensus-based FedAvg-inspired algorithm (referred to as CFA), supposing sparse connectivity. Consensus-driven FL (C-FL) \cite{21}, developed by Luca Barbieri et al. It is a decentralized, modular approach to learning FL which is suitable for Point Net compatible deep Systems and Lidia point cloud processing for road actor categorization. \cite{22} proposed fully decentralized paradigms driven by consensus methods on tumor segmentation. In addition, some Blockchain-based decentralization schemes have been proposed\cite{23,24}. Recently, \cite{25}, modeled the computation and communication resource in the FL by blockchain, and improved the utility between learning performance and resource consumption by controlling the number of local iterations in FL. The work in  \cite{26} shows the energy consumption model and optimizes the performance by allocating energy resources. The aforementioned methods avoid single-point failure. However, there are considerable costs associated with data storage and computing on the blockchain. On the other hand, uploading a whole machine learning model on the blockchain would be computationally burdensome and might result in significant lag.

All aforementioned works solve important problems. However, a key assumption is made in the studies: that data is independently and identically distributed (iid) over clients. However, the processing techniques of redundant data have not yet been widely applied and researched in decentralized federated learning. Thus, it is still a thorny problem to design a DFL method based on redundant data when considering system decentralization simultaneously.

\section{The Decentralized Federated learning}\label{sec3}

Decentralized FL approaches enable the sharing and synchronization of the local model parameters over networking with neighbors without relying on the servers. The proposed DFL approaches combine local models with neighboring ones by algorithms. After that, they update the models using local data and other’s parameters. When local models converge to satisfy a target loss or accuracy, the DFL process typically concludes after a number of communications rounds.

DFL aims to learn every local model $\hat{y}\left(W\right)$, with matrix $W$ encapsulates model parameters including Neural Networks weights $w_N$ and biases $b_N$ and $x$ is the input data. A deep Neural Network is composed of $N$ layers, including the input layer, hidden layers, and the output layer. The model iteratively computes a nonlinear function of a weighted sum of the input values in each layer. The function of the last layer is defined as:
\begin{equation}
\hat{y}\left(W^{(N)} ; x\right)=f_N\left(w_N^T h_{N-1}+b_N\right)
\end{equation}
Meanwhile, in other layers ($n=1,……N-1$) , the function is depicted in
\begin{equation}
f_n\left(w_n^T h_{n-1}+b_n\right)
\end{equation}
 and $h_{N-1}$ is an output of the last layer, while for $n=0$,we have $h_0=x$.
 In the DFL, the parameters $W$ can be learned by applying a minimization a global loss function $L(W)$:
\begin{equation}
\min _w L(W)=\min _w \sum_{k=1}^K \frac{E_k}{E} \times L_k(w)
\end{equation}

With $L_k$ being the local loss function observed by client $k$ and $E_k$ being the size of the $k$-th data set under the non-IID assumption. By alternately optimizing a local model at each client and engaging in a round of neighborhood communication to acquire an updated global model, DFL was able to tackle this problem iteratively. In general, the task initialization is implemented in each client at $t=0$. At iteration $t > 0$, each client sends local model parameters to their neighborhood. Then, every client updates the local model by aggregating the local model parameters and neighborhood parameters at $t-1$. This is solved by gradient methods, such as Stochastic Gradient Descent (SGD):

\begin{equation}
W_{t+1}=W_t-\mu_s \times \nabla L\left(W_t\right)
\end{equation}

Where $\mu_s$ is the learning rate of SGD, and $\nabla$ is the gradient of the loss by backpropagation. Considering the above described local model optimization and aggregation steps, training is repeated until each model converges to, or the desired training accuracy is obtained.

\section{THE METHOD}\label{sec4}

\subsection{Overall Design}\label{subsec4.1}
In this section, we propose C-DFL, a decentralized federated learning framework with a novel aggregation paradigm of local model updates to reduce the impact of redundant data on different local models. Specifically, in Section 4.2, we explore a novel redundant data processing solution by mapping data into a compact representation as a record of local data distribution.
In Section 4.3, we propose a 
decentralized federated learning framework, C-DFL, for distributed environment understanding of in-network vehicles.

The proposed algorithm designs the aggregation paradigm of local model updates by mapping data into a compact representation as a record of local data distribution. Through exchanging this type of compact representations, base stations can filter redundant data and thus speed up local updating and update aggregation. The C-DFL algorithm mainly includes two parts: the processing of redundant data and local model training. In the following, this paper details the redundant data processing and model training in section 4.2 and section 4.3.

\subsection{Redundant Data Processing}\label{subsec4.2}

In reality, the V2X network can generate a large amount of data. The data are collected by the camera on vehicles. Then vehicles send them to a nearby base station by V2X, which can lead to the base station having a lot of duplicate data captured by nearby vehicles. Thus, that can affect the accuracy and convergence speed of model training. Therefore, we design CND, a method that maps data into a compact representation. 

A detailed description of the CND is provided in Algorithm 1. On line 2, a new item is hashed, and the generated hash value. By different hash functions, we can obtain different bitmaps. To obtain the estimation result, we search the number of “1” in all the bitmaps (line 6) and calculate the arithmetic mean of these numbers, obtaining the carnality estimation of the dataset (line 9). Specifically, in the hash, we have considered features (separated by semicolon) in each item as tokens and assigned weights to features (lines 11-12). For generating an n-bit simple hash (line 15), we have used an n-bit Jenkin hash function. For each item, weighted all feature vectors, in accordance with the calculation rules, it encounters a hash value and weight, encountering 0 hash values and weight negative multiplication (line 16-20). The weighting result of the various feature vectors is accumulated, and there is only one sequence string (lines 24-25).

\begin{algorithm}[htbp]
\caption{Counting Non-repeated data (CND)}
\label{Algorithm 1}
\begin{algorithmic}[1] 
\REQUIRE ~~\\ 
$T$: a new item; \\
$Bitmap$ : vector of size $n$, initialized to $0$;
\ENSURE ~~\\ 
   avg: the estimation of cardinality of the items;
    \FOR{each $i \in [0,2]$}
     \WHILE {the sampling period is not end}
       \STATE hash$\gets hash_i(item)$;
       \STATE $Bitmap[hash]\gets 1$;
     \ENDWHILE 
     \STATE $count[i]$ $\gets$ Search the number of the 1 bit \\in the bitmap;
    \RETURN $count[i]$;
    \ENDFOR
    \STATE $avg$ $\gets$ $(count[0]+count[1]+count[2])/3$;
    \STATE begin $hash(item)$:
     \STATE $feature[i] \gets tokenize$;
     \STATE $weight[i]{\gets} $assign weights to features$ $;
     \STATE $v$ = vector of size $n$, initialized to $0$;
     \FOR{each feature $i$ in a item}
     \STATE $feature\_hash$=make $n$-bit hash(feature);
     \FOR {$i=1$ to $n$}
     \IF {$i^{th}\ bit\ of feature==1$}
     \STATE ${v[i]\gets v[i]+weight(feature)}$;
     \ELSE
     \STATE ${v[i]\gets v[i]-weight(feature)}$;
     \ENDIF
     \ENDFOR
     \ENDFOR
     \STATE $bit\_vector$=vector of size $n$, initialized to $0$
     \FOR{$i=1$ to $n$}
     \IF{$v[i]>0$} 
     \STATE$bit\_vector[i]=1$
     \ENDIF
     \ENDFOR
     \RETURN bit\_vector to decimal
\end{algorithmic}
\end{algorithm}
\subsection{Model Training}\label{subsec4.3}
This section mainly introduces the algorithm for model training, and the proposed method allows the base station to rely on cooperation with neighbor base stations and local intranet processing to learn model parameters. We construct a ring network topology $G=(V,E)$ with the set of base stations $V={1,2,...,k}$ and edges (links) $E$. The $K$ distributed base stations are connected through a decentralized communication architecture based on V2X communications. The neighbor set of base station $k$ is denoted as $N_{\overline{k}}$ , with cardinality$ |N_{\overline{k}}|$. Notice that we include base station $k$ in the $N_k$ while ${N_{\overline{k}}}= N_k\backslash \{k\}$ does not. As introduced in the previous section, each base station has a database $E_k$ of examples $(x_h, y_h)$ that are used to train a local NN model $W_{t,k}$ at some epoch $t$. The model maps input features $x$ into outputs $y^{\prime}(W_{t,k},x)$  as in (1). A cost function, generally non-convex, as $L_k(W_{t,k})$ in (3), is used to optimize the weights $W_{t,k}$ of the local model.
\begin{algorithm}[htbp]
\caption{C-DFL}
\label{Algorithm 2}
\begin{algorithmic}[1] 
\REQUIRE ~~\\ 
$T$: Dataset;
\ENSURE ~~\\ 
   Category;
    \STATE initialize $w_{0, k} \leftarrow $ base station $k$
    \FOR{round $t \in [1,n]$}
     \STATE receive $w_{t,i}\{i \in N_k\}$,  bitmaps
     \STATE $\varphi_{t,k}\gets w_{t,k}$
     \FOR{all base stations $i \in N_k$}
     \STATE$\varphi_{t, k} \leftarrow \eta_{i, i} \varphi_{t, k}+\gamma_{t} \eta_{k, i}\left(w_{t, i}-w_{t, k}\right)$
     \ENDFOR
     \STATE$w_{t+1, k}=\text { ModelUpdate }\left(\varphi_{t, k}\right)$
     \STATE send ($W_{t+1, k}$), bitmaps
   
    \STATE begin $\text { ModelUpdata }\left(\varphi_{\mathrm{t}, \mathrm{k}}\right)$
    \STATE $B \leftarrow \text { mini-batches of size } \boldsymbol{B}$

    \vspace{1ex} 
    \STATE $m_{t+1, i} \leftarrow \beta_{1} m_{t, i}+\left(1-\beta_{1}\right) \nabla L_{t, i}\left(\psi_{t, i}\right)$
    \vspace{1ex}
    \STATE $v_{t+1, i} \leftarrow \beta_{2} v_{t, i}+\left(1-\beta_{2}\right) \nabla^{2} L_{t, i}\left(\psi_{t, i}\right)$
   \vspace{1.5ex}
    \STATE $\psi_{t+1, i} \leftarrow \psi_{t, i}-\frac{\sqrt{1-\beta_{2}^{t}}}{1-\beta_{1}^{t}} \cdot \frac{m_{t+1, i}}{\sqrt{v_{t+1, i}}+\sigma}$
     \ENDFOR
     \STATE ${w}_{t, k} \leftarrow \phi_{t, k}$
     \RETURN (${w}_{t, k}$)
     \STATE end $\text { ModelUpdata }\left(\varphi_{\mathrm{t}, \mathrm{k}}\right)$
\end{algorithmic}
\end{algorithm}

First, initialize the parameters $w_{0,k}$ and compute the bitmaps of local data by CND (lines 2-5) at time $t = 0$ for each base station. 
After a certain round of local training time $t>0$, base station $k$ sends its model updates $w_{t,k}$ and bitmaps of the dataset to its neighbor by the V2X network. 
Meanwhile other base stations $k$ receive weights and bitmaps from neighbors $w_{t,i}$, $i\in N_{k}$.
On the received bitmaps, the base station $k$ uses the local data set to continue to calculate the hash value by CND.
After the above processing, the base station $k$ calculates the number of different data between it and other neighbor base stations.
Then, it obtains the weight of the model aggregated based on the number of different data.
Finally, it updates its model $w_{t,k}$ at time $t$.
\begin{equation}
\varphi_{t, k}=\eta_{i, i} \varphi_{t, k}+\gamma_t \sum_{i \in N_{k}} \eta_{k, i}\left(W_{t, i}-W_{t, k}\right)
\end{equation}
Where $\gamma_t$ is the consensus step size and $\eta_{k,i}$ is the mixing weights for the models which are stetted as:
\begin{equation}
\eta_{k, i}=\frac{\dddot{E}_i}{\sum_{k \in N_{\overline{i}}} \dddot{E_k}}
\end{equation}

\begin{equation}
\dddot{E}_k=\frac{E_{k^\prime}}{E_{k}}
\end{equation}

where $E_k$ is the size of the dataset on base station $k$ and $E_{k^\prime}$ is the number of the dataset processed by the CND, where the mixing weights  $\eta_{k, i}$ are adapted on each epoch $t$ based on current validation accuracy or loss metrics. The consensus step size $\gamma_t$ can be chosen as $\gamma_t\in(0,1/\nabla)$, and
$
\nabla=\max _k\left(\sum_{i \in N_k} \eta_{t, i}\right)
$.

Once the consensus process is completed, the base station $i$ sends the parameters $w_{t,k}$ and runs a local optimizer to minimize local loss $L$ in (3). Considering Adam is an optimizer, this last stage is implemented as $
W_{t+1}=W_t-\mu_s \times \nabla L\left(\mathrm{~W}_t\right)
$ with:

\begin{equation}
\left\{\begin{array}{l}
\Delta \psi_{t, i}=\mu_t \cdot \frac{\sqrt{1-\beta_2^t}}{1-\beta_1^t} \cdot \frac{m_{t+1, i}}{\sqrt{v_{t+1, i}}+\delta} \vspace{2ex} \\
m_{t+1, i}=\beta_1 m_{t, i}+\left(1-\beta_1\right) \nabla L_{t, i}\left(\psi_{t, i}\right) \vspace{2ex}\\
v_{t+1, i}=\beta_2 v_{t, i}+\left(1-\beta_2\right) \nabla^2 L_{t, i}\left(\psi_{t, i}\right)
\end{array}\right.
\end{equation}

Where $m_{t+1, i}$ is the estimates of first moment of the gradients $\nabla L_{t, i}\left(\psi_{t, i}\right)$ at round $t+1$, $v_{t+1, i}$ is the estimates of second moment. $\beta_1$ and $\beta_2$ are  decaying averages. $\delta$ is a small value for numerical stability. A model update in (8) is computed over a mini-batch $B (B<E_{k})$ of local training. After the model update, the new parameters $W_{t+1,k}$ are forwarded to the neighbors of base station $k$, and a new round starts. The model is iterated through this process until the appropriate loss values are reached.

\section{EXPERIMENTAL EVALUATION}\label{sec5}

We report experiments conducted on two real-world datasets, which are: MINIST\cite{29} and BIRD-400\cite{30} to validate our proposed C-DFL approach.

\subsection{Simulation Setup}\label{subsec5.1}
In order to validate our method, we simulate the experiment with MLP and VGG on Sim4DistrDL\cite{27}. We choose the common edge network topology (as depicted in Fig.4). The topology used in the simulation process is a ring topology. It includes four connected base stations and their adjacent end vehicles.
The connection between vehicles and base stations via V2X. The ML model's parameters are exchanged through V2X communications. The base stations and vehicles are connected through wireless links. The vehicles are used for capturing data, while base stations are used for training models. The data required for model training are all released by the nearby vehicles. The base station uses the received parameters and the data received by vehicles to train DNN models in a collaborative way.
\begin{figure*}[h!]
\centering
\includegraphics[width=0.8\linewidth]{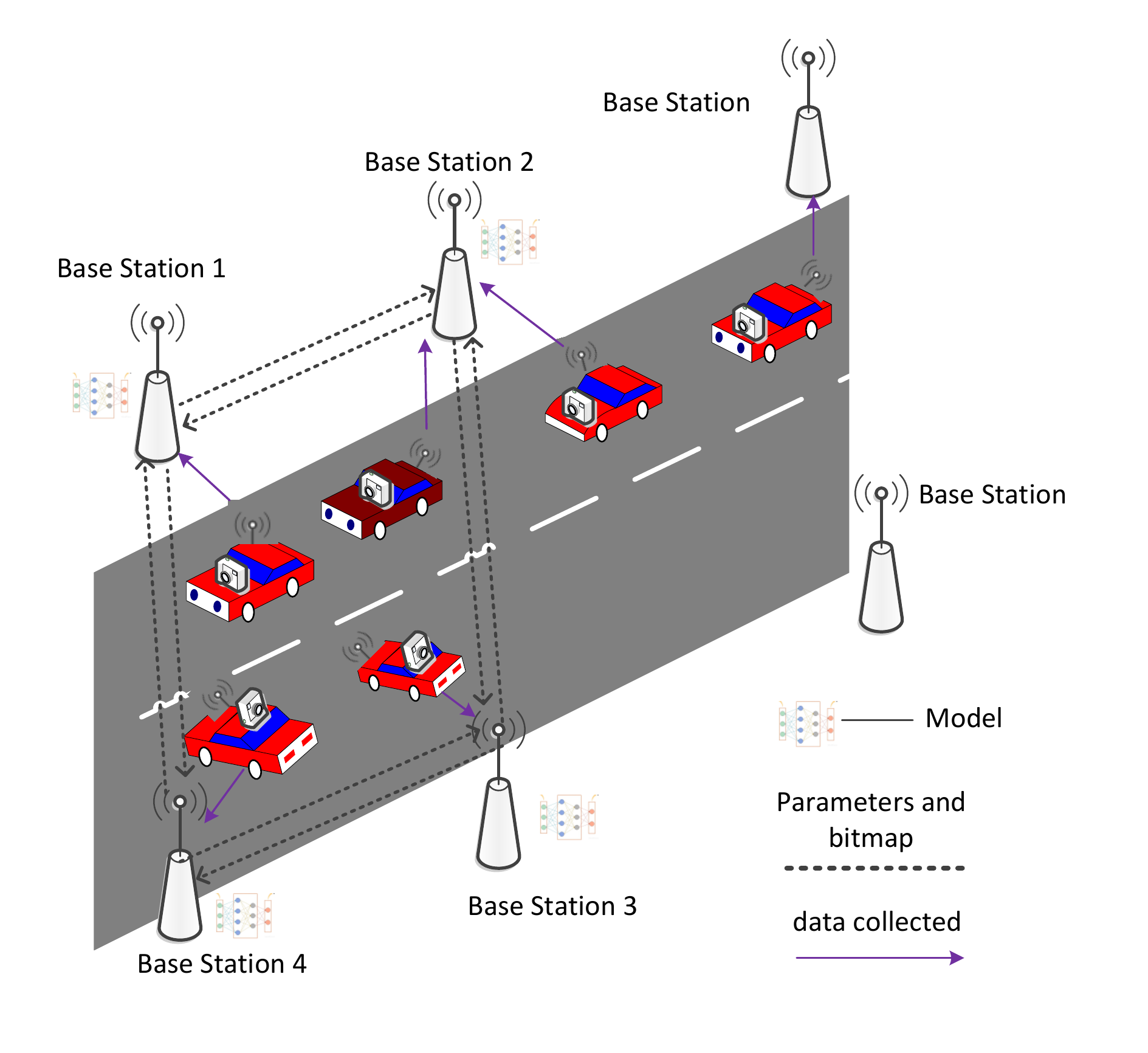}
\caption{Simulation Topology of the DFL vehicular scenario}
\label{fig3}
\end{figure*}
\begin{figure}[h!]
\centering
\includegraphics[width=0.8\linewidth]{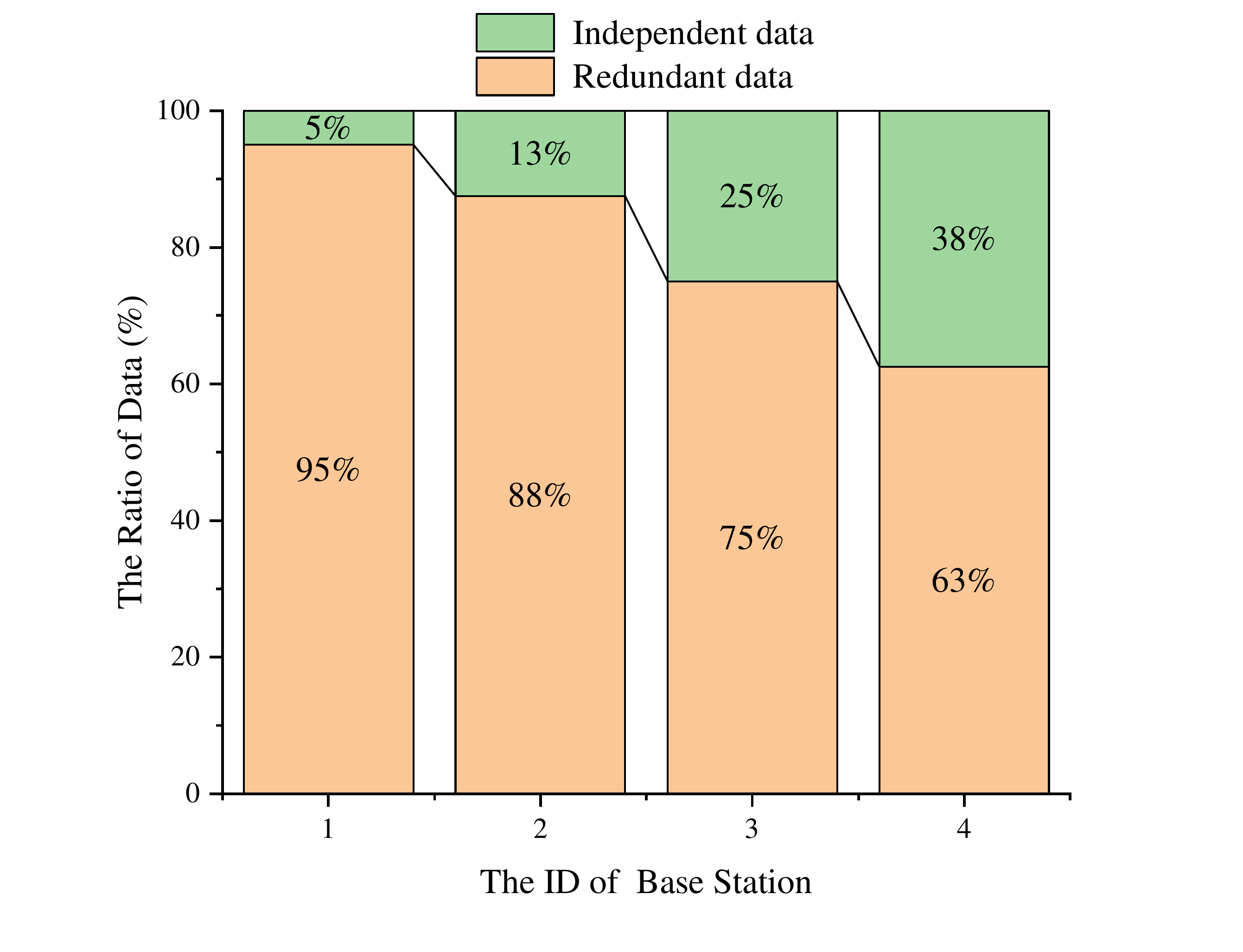}
\caption{The Data Distribution on MLP}
\label{fig4}
\end{figure}

\begin{figure*}[h!]
\begin{center}
\includegraphics[width=0.7\linewidth]{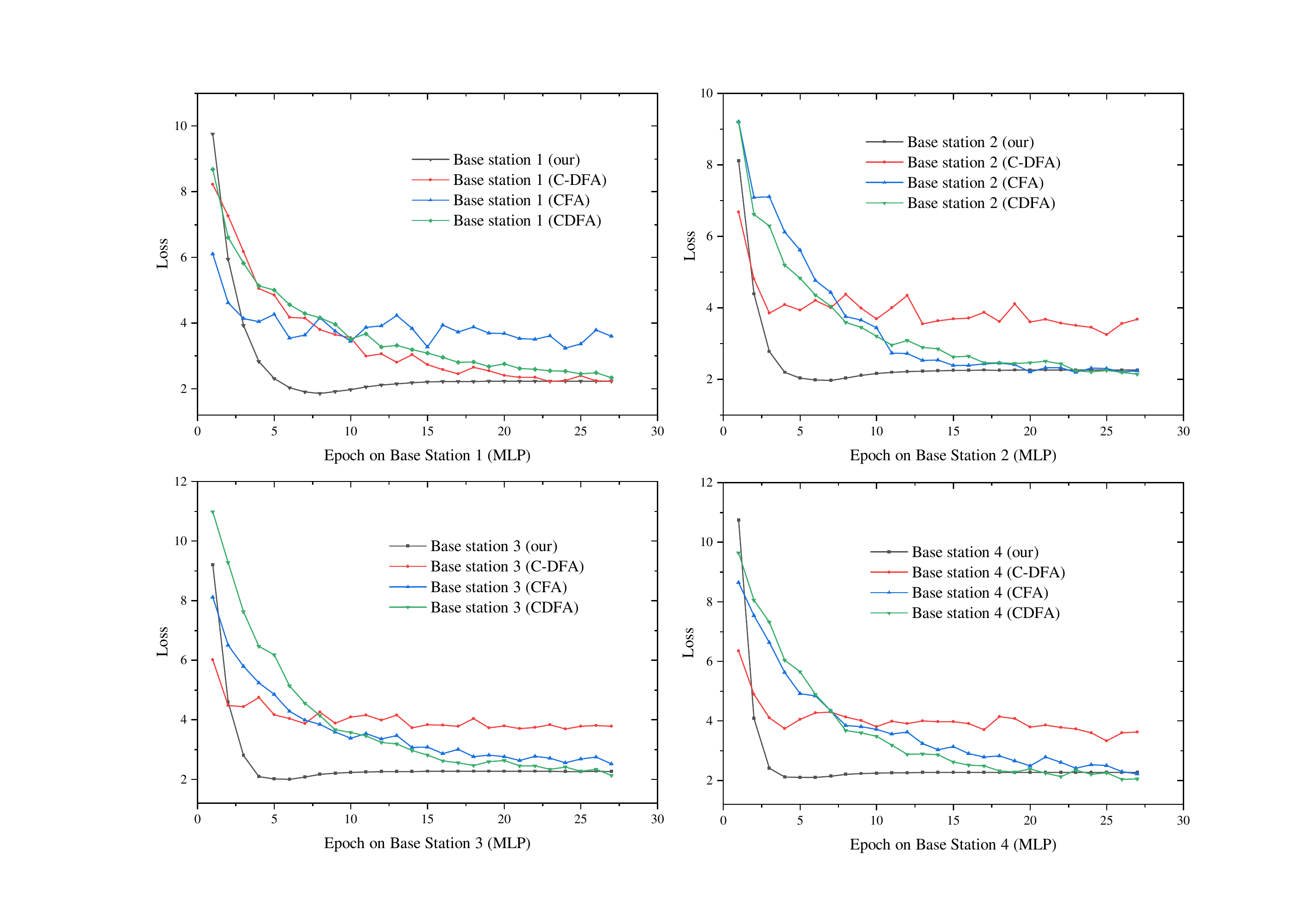}
\caption{The Cross-Entropy Loss of MLP}
\label{fig5}
\end{center}
\end{figure*}

\begin{figure*}[h!]
\begin{center}
\includegraphics[width=0.7\linewidth]{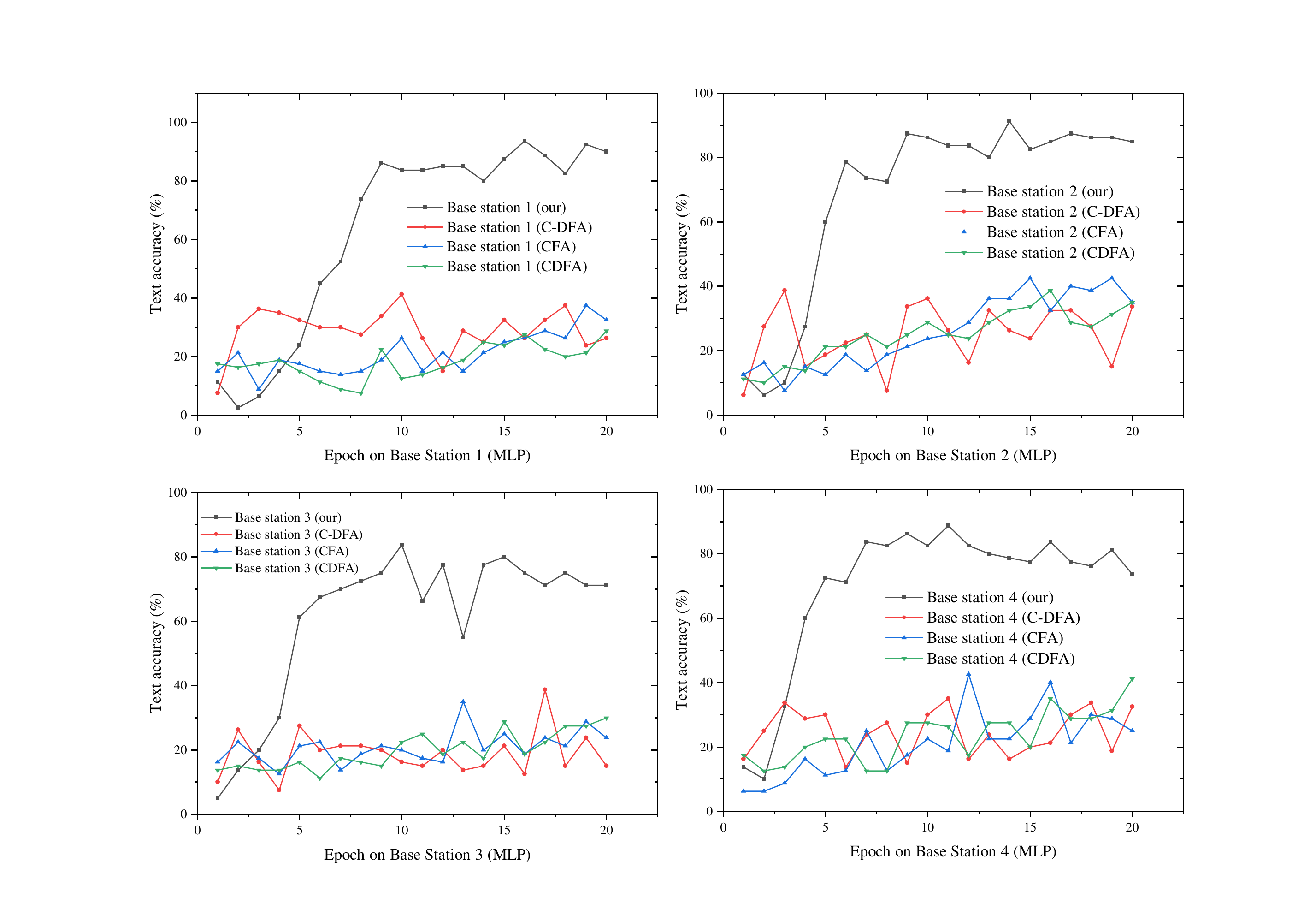}
\caption{Test accuracy (\%) comparison of  C-DFL to CFA and CDFA on MINIST (MLP)}
\label{fig6}
\end{center}
\end{figure*}
\begin{figure}[h!]
\centering
\includegraphics[width=0.9\linewidth]{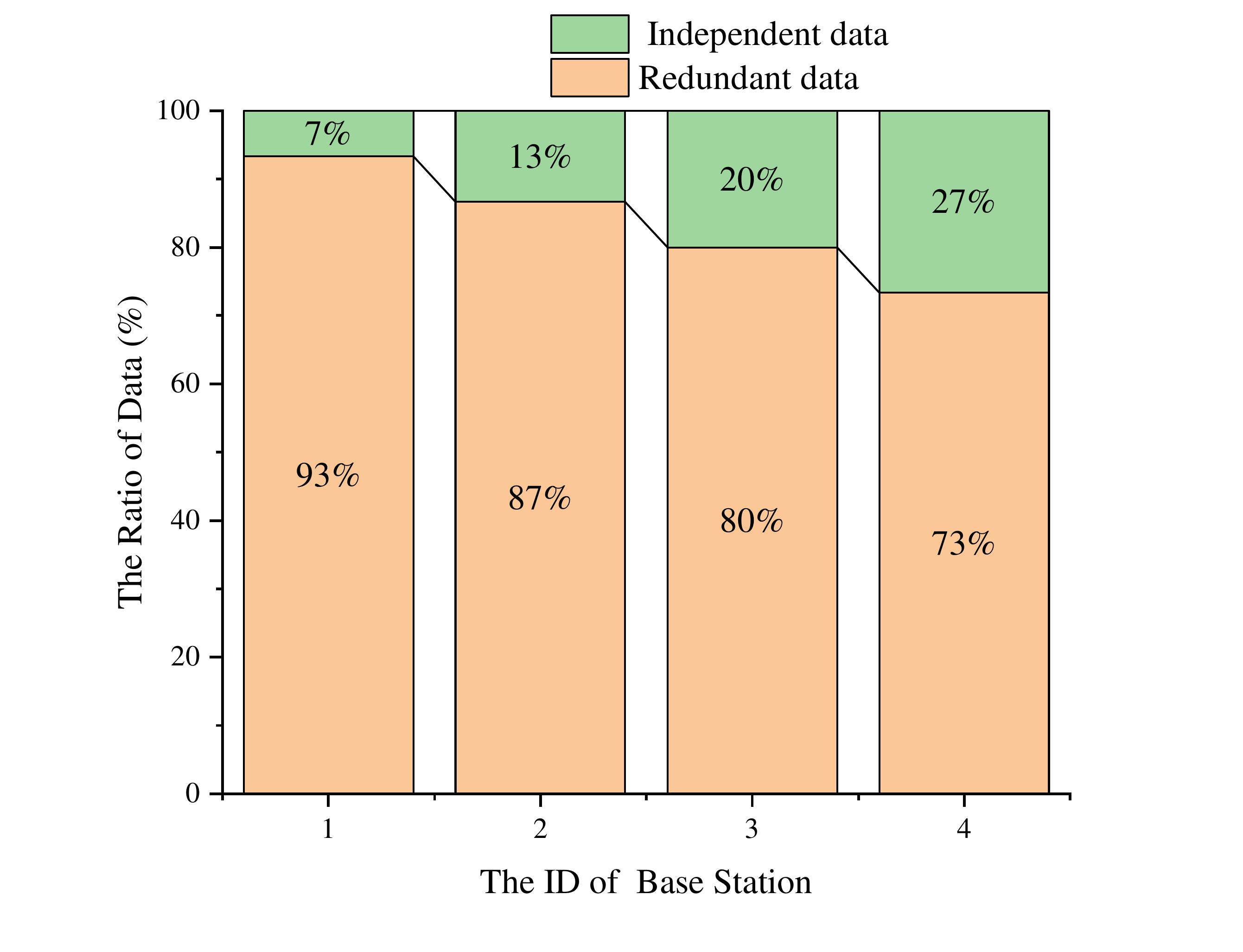}
\caption{The Data Distribution on VGG}
\label{fig7}
\end{figure}

\begin{figure*}[h!]
\begin{center}
\includegraphics[width=0.7\linewidth]{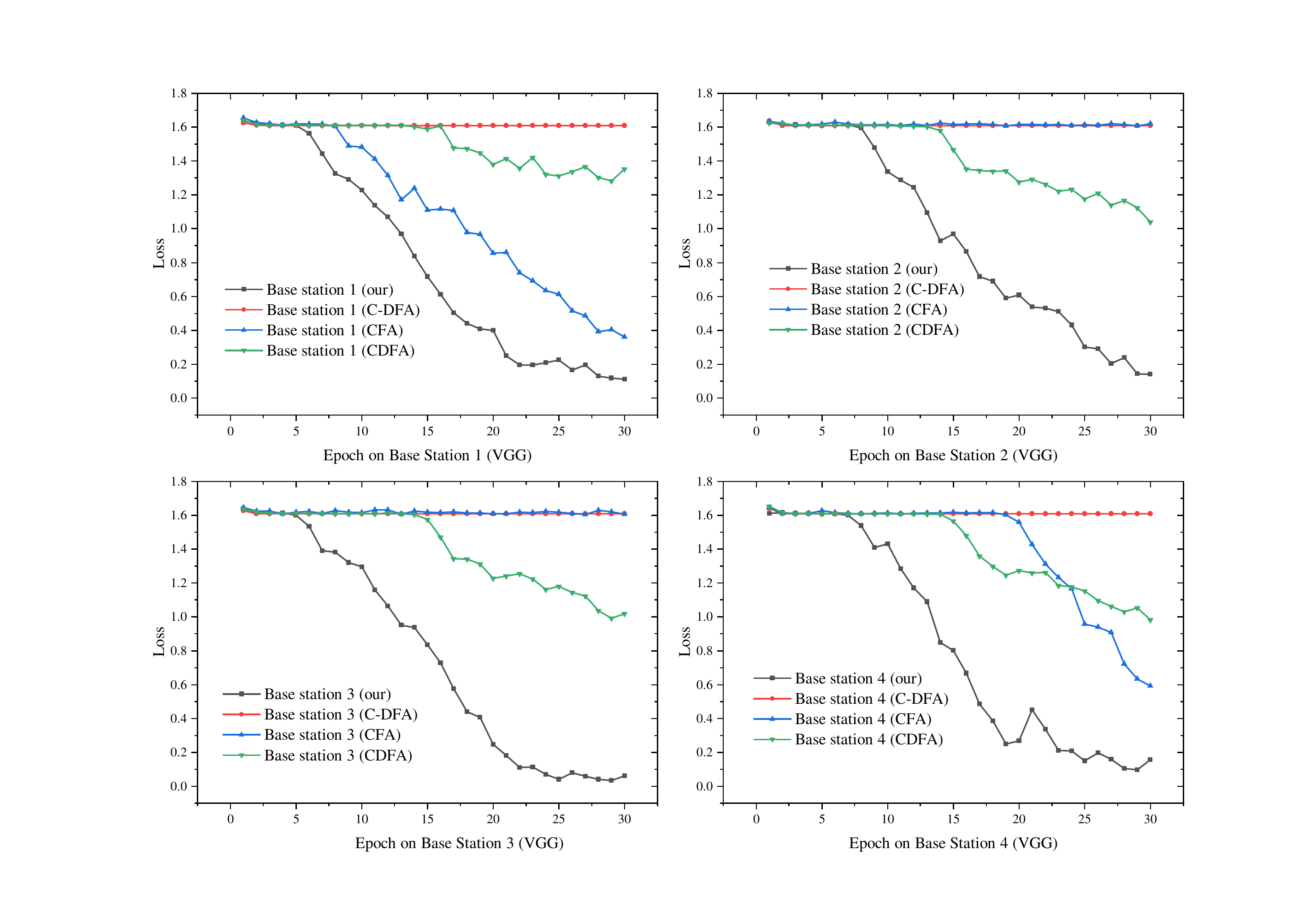}
\caption{The Cross-Entropy Loss of CNN (VGG)}
\label{fig8}
\end{center}
\end{figure*}

\begin{figure*}[h!]
\begin{center}
\includegraphics[width=0.7\linewidth]{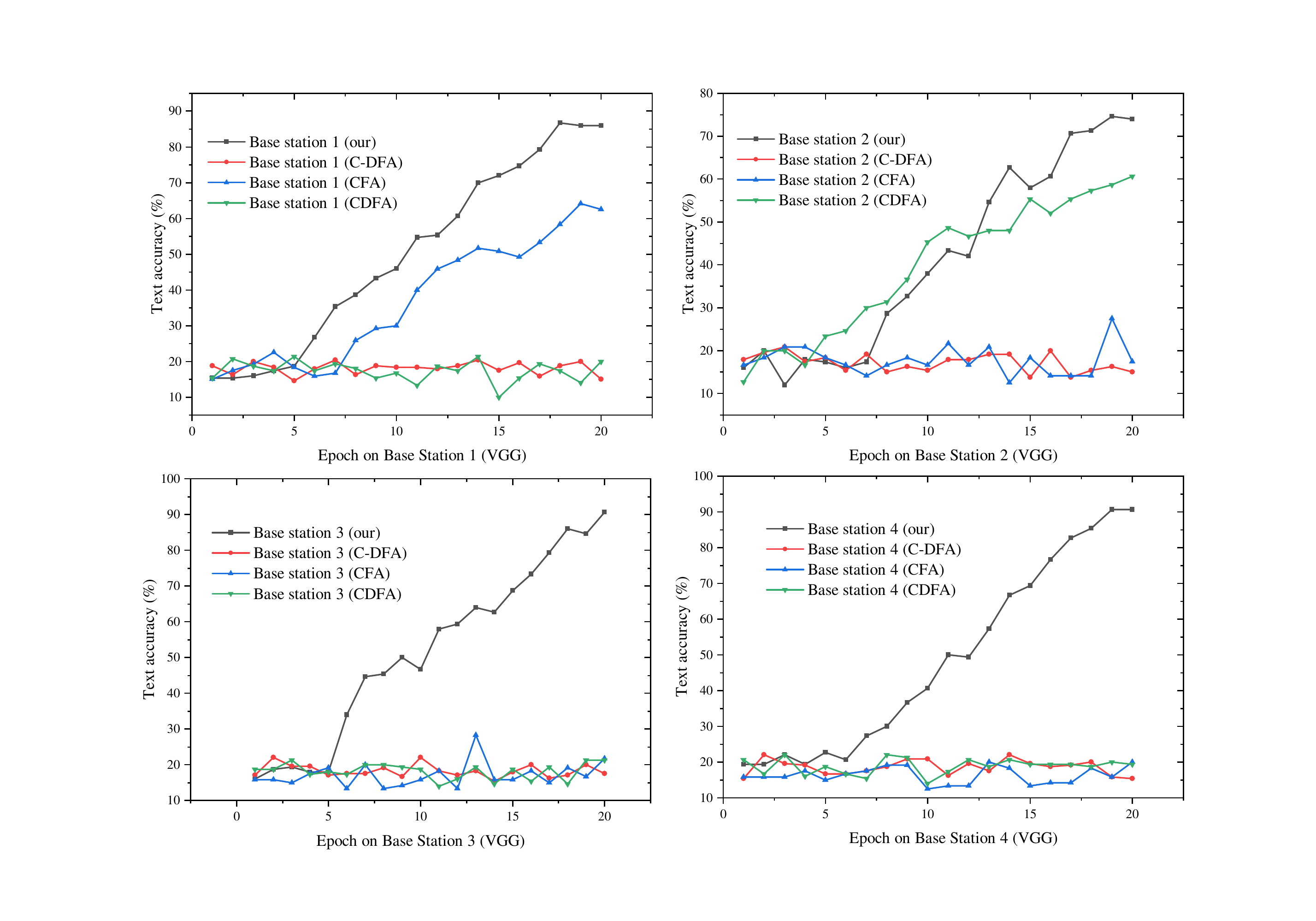}
\caption{Test accuracy (\%) comparison of C-DFL to CFA and CDFA on BRID-400 (VGG)}
\label{fig9}
\end{center}
\end{figure*}

As can be seen from the above topology diagram, the neighbor of 4-edge base station sets consists of ${N_{\overline{1}}}= \{4,2\}$, ${N_{\overline{2}}}= \{1, 3\}$, ${N_{\overline{3}}}= \{2, 4\}$, and ${N_{\overline{4}}}= \{3,1\}$. Each $k$th base station has a database of $E_k$ local training data.
\subsection{Datasets}\label{sec5.2}
Our experiments are performed on two datasets for classification, MNIST and BIRDS 400. The MNIST consists of 10 classes (from 0 to 9) with the dimension of 786 ($28 X 28 X 1$), respectively. BIRD consists of 400 bird species, and all images are $224 X 224 X 3$ color images in jpg format.

\subsection{Baseline Models}\label{sec5.3}
We consider three baseline algorithms: (1) Consensus-Based FA (CFA)\cite{13}, (2) Consensus-driven federated averaging (C-DFA)\cite{21} (3) Consensus-Driven FA (CDFA) \cite{7}. The training of the above algorithms is based on decentralized federated learning. Specifically, the C-DFA algorithm is implemented by applying the federated optimization to a variable number $Q \leq N$ of layers (FC and CN layers) in the NN. More specifically, performances are analyzed by varying the fraction $M = Q /N$ of the layers subject to the C-FL process. We consider C-DFA $M = 100\%$ as a baseline. For CDFA, we don’t consider encoding and decoding, and use it as another comparison scheme.

\subsection{Experimental Results}\label{sec5.4}

In this section, we give the numerical results for the evaluation of the proposed C-DFL. The method is evaluated in terms of loss and accuracy using the previously mentioned network topology configuration. In the following, Sec. 5.4.1 details the results of the  MLP model based on MINIST. Sec. 5.4.2 describes the results of the VGG model based on BRID-400 and illustrates the more accurate comparison of the two models. 

\subsubsection{MINIST}\label{sec5.4.1}
The MNIST consists of 10 classes (from 0 to 9) of signals with a dimension of 786 (28x28x1). In this work, we consider solving a $10$-label classification problem with a data federation formed between different vehicles. In our setting, each base station collects $320$ training samples (including $10$ categories) and $80$ testing samples (see fig4 for the data distribution). Each base station utilizes local neural networks that have one hidden layer with 30 units to learn input samples. We set $learning$  $rate = 0.0001$, $\beta_1 = 0.9$, $\beta_2 = 0.999$, $\delta  = 1e-7$ and $batch$ $size = 32$. Each base station sends and aggregates the learned weights and biases of neural networks with its neighbors via the V2X network.

Fig. 5 reports the loss of C-DFL (our method) and baselines on each base station, while Fig. 6 presents the corresponding test accuracy for our method and baselines. It can be seen from Fig. 5 that the baseline algorithms converge very slowly due to redundant data. On the other hand, our method can provide highly accurate results outperforming baselines. It outperforms Consensus-Driven FA \cite{25} and CDFA significantly and outperforms CFA by a large margin (see Figure 6). 

\subsubsection{BIRD-400}\label{sec5.4.2}

This Dataset contains $58388$ training images and $2000$ test images of $400$ categories with a shape of $224 X 224 X 3$. This is an extremely high-quality dataset where each image only contains one bird, and the bird usually occupies at least $50\%$ of the image's pixels. Each base station has 120 training samples (including 5 categories and redundant data) and 30 testing samples (the data distribution as depicted in Fig7). Each base station uses one local Visual Geometry Group (VGG) model to learn from input image samples. We set $learning$ $rate$ = $0.001$, $\beta_1 = 0.9$, $\beta_2 = 0.99$, $\delta  = 1e-7$ and $batch$ $size$ = $10$. After training at local, each base station sends and aggregates the learned weights and biases of the model with the in-network vehicles.
From Fig.8 and Fig.9, we have the following observations. Our scheme is obviously faster than CDFA and C-DFA in terms of convergence speed and also has some improvement over CFA. This suggests that, the baseline algorithms are poorly adapted to redundant data.

Table1-4 shows the training rounds to achieve a specific accuracy(about $80$ percent) in terms of different algorithms. C-DFL outperforms the baselines in the data processing period. This demonstrates that filtering redundant data brings great benefits for improving the accuracy and convergence speed of federated learning models. Our method achieves $80$ percent accuracy in about 8 epochs, whereas other methods only achieve $50$ accuracy in 100 epochs. While using the VGG model, C-DFL reduceds convergence delay by half compared with C-DFA and CDFA. In addition, compared with the CFA algorithm, our scheme also has a significant improvement. Results also indicate that, the reduction process of the impact of redundant data at local is essential to improve aggregation performance among the base stations. 

\begin{table}[htbp]
\renewcommand\arraystretch{1.3}
\centering
\caption{The Epoch and the Test accuracy}
\label{table1}
\setlength{\tabcolsep}{1.2mm}
\begin{tabular}{ccccc}
\hline
          & \begin{tabular}[c]{@{}c@{}}Base station\\1 (Our)\end{tabular} & \begin{tabular}[c]{@{}c@{}}Base station\\1 (CFA)\end{tabular} & \begin{tabular}[c]{@{}c@{}}Base station\\1 (C-DFA)\end{tabular} & \begin{tabular}[c]{@{}c@{}}Base station\\1 (CDFA)\end{tabular} \\ \hline
MLP       & \color[HTML]{FF0000}\textbf{9(0.86)}      & 100(0.54)         & 100 (0.45)     & 100(0.44)     \\
CNN (VGG) & \color[HTML]{FF0000}\textbf{19(0.86)}     & \textbf{19(0.86)} & 27(0.81)       & 47(0.81)      \\ \bottomrule
\end{tabular}
\end{table}
\begin{table}[htbp]
\renewcommand\arraystretch{1.3}
\centering
\caption{The Epoch and the Test accuracy}
\label{table2}
\setlength{\tabcolsep}{1.2mm}
\begin{tabular}{ccccc}
\hline
          & \begin{tabular}[c]{@{}c@{}}Base station\\2 (Our)\end{tabular} & \begin{tabular}[c]{@{}c@{}}Base station\\2 (CFA)\end{tabular} & \begin{tabular}[c]{@{}c@{}}Base station\\2 (C-DFA)\end{tabular} & \begin{tabular}[c]{@{}c@{}}Base station\\2 (CDFA)\end{tabular} \\ \hline
MLP       & \color[HTML]{FF0000}\textbf{9(0.88)}      & 100(0.58)         & 100 (0.31)     & 100(0.45)     \\
CNN (VGG) & \color[HTML]{FF0000}\textbf{18(0.86)}     & 28(0.85) & 52(0.82)       & 43(0.80)      \\ \bottomrule
\end{tabular}
\end{table}
\begin{table}[htbp]
\renewcommand\arraystretch{1.3}
\centering
\caption{The Epoch and the Test accuracy}
\label{table3}
\setlength{\tabcolsep}{1.2mm}
\begin{tabular}{ccccc}
\hline
          & \begin{tabular}[c]{@{}c@{}}Base station\\3 (Our)\end{tabular} & \begin{tabular}[c]{@{}c@{}}Base station\\3 (CFA)\end{tabular} & \begin{tabular}[c]{@{}c@{}}Base station\\3 (C-DFA)\end{tabular} & \begin{tabular}[c]{@{}c@{}}Base station\\3 (CDFA)\end{tabular} \\ \hline
MLP       & \color[HTML]{FF0000}\textbf{10(0.84)} & 100(0.4)                                 & 100 (0.15)     & 100(0.44)     \\
CNN (VGG) & 18(0.87)          & {\color[HTML]{FF0000} \textbf{15}}\textbf{(0.86)} & 37(0.8)        & 43(0.82)      \\ \bottomrule
\end{tabular}
\end{table}

\begin{table}[htbp]
\renewcommand\arraystretch{1.3}
\centering
\caption{The Epoch and the Test accuracy}
\label{table4}
\setlength{\tabcolsep}{1.2mm}
\begin{tabular}{ccccc}
\hline
          & \begin{tabular}[c]{@{}c@{}}Base station\\4 (Our)\end{tabular} & \begin{tabular}[c]{@{}c@{}}Base station\\4 (CFA)\end{tabular} & \begin{tabular}[c]{@{}c@{}}Base station\\4 (C-DFA)\end{tabular} & \begin{tabular}[c]{@{}c@{}}Base station\\4 (CDFA)\end{tabular} \\ \hline
MLP       & {\color[HTML]{FF0000} \textbf{7(0.84)}}                 & 100(0.58)                                               & 100(0.6)                                                  & 100(0.61)                                                \\
CNN (VGG) & {\color[HTML]{FF0000} \textbf{17(0.83)}}                & {\color[HTML]{000000} 19(0.86)}                         & 30(0.83)                                                  & 47(0.87)                                                 \\ \hline
\end{tabular}
\end{table}

\section{Conclusion}\label{sec6}

We addressed a novel idea of decentralized data processing with a federated learning framework based on consensus in vehicles. To satisfy concerns of edge-cloud cooperation for new intelligent transport scenarios, we explore a decentralized aggregation paradigm of local model updates. Extensive simulations on NS-3 demonstrate that through efficient cooperation at the edge.
The evaluation has also shown that this is useful to reduce the learning time and meet the challenging requirements foreseen for full self-driving scenarios.
Finally, although this paper addressed a classification task as an application of connected vehicles, the proposed framework can be generalized to perform a wider range of tasks.

\section*{Acknowledgments}
This work was supported by the Open Foundation of State Key Laboratory of Networking and Switching Technology (Beijing University of Posts and Telecommunications) (SKLNST-2020-1-18), the National Science Foundation of China (61962045, 62062055, 61902382, 61972381), the Science and Technology Planning Project of Inner Mongolia Autonomous Region (2019GG372), the Science Research Project of Inner Mongolia University of Technology (BS201934). 
\medskip


\begin{thebibliography}{29}
\providecommand{\natexlab}[1]{#1}
\providecommand{\url}[1]{\texttt{#1}}
\expandafter\ifx\csname urlstyle\endcsname\relax
  \providecommand{\doi}[1]{doi: #1}\else
  \providecommand{\doi}{doi: \begingroup \urlstyle{rm}\Url}\fi

\bibitem[Yurtsever et~al.(2019)Yurtsever, Lambert, Carballo, and Takeda]{1}
E.~Yurtsever, J.~Lambert, A.~Carballo, and K.~Takeda.
\newblock A survey of autonomous driving: Common practices and emerging
  technologies.
\newblock 2019.

\bibitem[Yang et~al.(2021)Yang, Cao, Xiong, Yuen, and Han]{2}
B.~Yang, X.~Cao, K.~Xiong, C.~Yuen, and Z.~Han.
\newblock Edge intelligence for autonomous driving in 6g wireless system:
  Design challenges and solutions.
\newblock \emph{IEEE Wireless Communications}, 28\penalty0 (2):\penalty0
  40--47, 2021.

\bibitem[Eskandarian et~al.(2019)Eskandarian, Wu, and Sun]{3}
A.~Eskandarian, C.~Wu, and C.~Sun.
\newblock Research advances and challenges of autonomous and connected ground
  vehicles.
\newblock \emph{IEEE Transactions on Intelligent Transportation Systems},
  PP\penalty0 (99):\penalty0 1--29, 2019.

\bibitem[Barbieri et~al.(2021)Barbieri, Savazzi, Brambilla, and Nicoli]{4}
L.~Barbieri, S.~Savazzi, M.~Brambilla, and M.~Nicoli.
\newblock Decentralized federated learning for extended sensing in 6g connected
  vehicles.
\newblock \emph{Vehicular Communications}, page 100396, 2021.

\bibitem[Liu et~al.(2021{\natexlab{a}})Liu, Chen, and Zhang]{5}
W.~Liu, L.~Chen, and W.~Zhang.
\newblock Decentralized federated learning: Balancing communication and
  computing costs.
\newblock 2021{\natexlab{a}}.

\bibitem[Xie et~al.(2019)Xie, Ma, Wang, Li, and Li]{6}
X.~Xie, L.~Ma, H.~Wang, Y.~Li, and X.~Li.
\newblock Diffchaser: Detecting disagreements for deep neural networks.
\newblock In \emph{Twenty-Eighth International Joint Conference on Artificial
  Intelligence {IJCAI-19}}, 2019.

\bibitem[Lian et~al.(2017)Lian, Zhang, Zhang, Hsieh, Zhang, and Liu]{7}
X.~Lian, C.~Zhang, H.~Zhang, C.~J. Hsieh, W.~Zhang, and J.~Liu.
\newblock Can decentralized algorithms outperform centralized algorithms? a
  case study for decentralized parallel stochastic gradient descent.
\newblock 2017.

\bibitem[Bonawitz et~al.(2019)Bonawitz, Eichner, Grieskamp, Huba, Ingerman,
  Ivanov, Kiddon, Konen, Mazzocchi, and Mcmahan]{8}
K.~Bonawitz, H.~Eichner, W.~Grieskamp, D.~Huba, A.~Ingerman, V.~Ivanov,
  C.~Kiddon, Jakub Konen, S.~Mazzocchi, and H.~B. Mcmahan.
\newblock Towards federated learning at scale: System design.
\newblock 2019.

\bibitem[Chen et~al.(2020)Chen, Tian, Liao, and Yu]{9}
Zheyi Chen, Pu~Tian, Weixian Liao, and Wei Yu.
\newblock Zero knowledge clustering based adversarial mitigation in
  heterogeneous federated learning.
\newblock \emph{IEEE Transactions on Network Science and Engineering},
  PP\penalty0 (99):\penalty0 1--1, 2020.

\bibitem[Lan et~al.(2017)Lan, Lee, and Zhou]{10}
G.~Lan, S.~Lee, and Y.~Zhou.
\newblock Communication-efficient algorithms for decentralized and stochastic
  optimization.
\newblock \emph{arXiv preprint arXiv:1701.03961}, 2017.

\bibitem[Sirb and Ye(2016)]{11}
B.~Sirb and X.~Ye.
\newblock Consensus optimization with delayed and stochastic gradients on
  decentralized networks. inbig data (big data), 2016 ieee international
  conference on.
\newblock pages 76--85, 2016.

\bibitem[Hardy et~al.(2018)Hardy, Le~Merrer, and Sericola]{12}
Corentin Hardy, Erwan Le~Merrer, and Bruno Sericola.
\newblock Gossiping gans: Position paper.
\newblock In \emph{Proceedings of the Second Workshop on Distributed
  Infrastructures for Deep Learning}, pages 25--28, 2018.

\bibitem[Lalitha et~al.(2018)Lalitha, Shekhar, Javidi, and Koushanfar]{13}
Anusha Lalitha, Shubhanshu Shekhar, Tara Javidi, and Farinaz Koushanfar.
\newblock Fully decentralized federated learning.
\newblock In \emph{Third workshop on Bayesian Deep Learning (NeurIPS)}, 2018.

\bibitem[Roy et~al.(2019)Roy, Siddiqui, P{\"o}lsterl, Navab, and Wachinger]{14}
Abhijit~Guha Roy, Shayan Siddiqui, Sebastian P{\"o}lsterl, Nassir Navab, and
  Christian Wachinger.
\newblock Braintorrent: A peer-to-peer environment for decentralized federated
  learning.
\newblock \emph{arXiv preprint arXiv:1905.06731}, 2019.

\bibitem[Hu et~al.(2019)Hu, Jiang, and Wang]{15}
Chenghao Hu, Jingyan Jiang, and Zhi Wang.
\newblock Decentralized federated learning: A segmented gossip approach.
\newblock \emph{arXiv preprint arXiv:1908.07782}, 2019.

\bibitem[Li et~al.(2020{\natexlab{a}})Li, Chen, Liu, Huang, Zheng, and Yan]{16}
Yuzheng Li, Chuan Chen, Nan Liu, Huawei Huang, Zibin Zheng, and Qiang Yan.
\newblock A blockchain-based decentralized federated learning framework with
  committee consensus.
\newblock \emph{IEEE Network}, 35\penalty0 (1):\penalty0 234--241,
  2020{\natexlab{a}}.

\bibitem[Qu et~al.(2020)Qu, Pokhrel, Garg, Gao, and Xiang]{17}
Youyang Qu, Shiva~Raj Pokhrel, Sahil Garg, Longxiang Gao, and Yong Xiang.
\newblock A blockchained federated learning framework for cognitive computing
  in industry 4.0 networks.
\newblock \emph{IEEE Transactions on Industrial Informatics}, 17\penalty0
  (4):\penalty0 2964--2973, 2020.

\bibitem[Lu et~al.(2020)Lu, Huang, Zhang, Maharjan, and Zhang]{18}
Yunlong Lu, Xiaohong Huang, Ke~Zhang, Sabita Maharjan, and Yan Zhang.
\newblock Blockchain and federated learning for 5g beyond.
\newblock \emph{Ieee Network}, 35\penalty0 (1):\penalty0 219--225, 2020.

\bibitem[Pokhrel(2021)]{19}
Shiva~Raj Pokhrel.
\newblock Blockchain brings trust to collaborative drones and leo satellites:
  An intelligent decentralized learning in the space.
\newblock \emph{IEEE sensors journal}, 21\penalty0 (22):\penalty0 25331--25339,
  2021.

\bibitem[Tedeschini et~al.(2022)Tedeschini, Savazzi, Stoklasa, Barbieri,
  Stathopoulos, Nicoli, and Serio]{20}
Bernardo~Camajori Tedeschini, Stefano Savazzi, Roman Stoklasa, Luca Barbieri,
  Ioannis Stathopoulos, Monica Nicoli, and Luigi Serio.
\newblock Decentralized federated learning for healthcare networks: A case
  study on tumor segmentation.
\newblock \emph{IEEE Access}, 10:\penalty0 8693--8708, 2022.

\bibitem[Barbieri et~al.(2022)Barbieri, Savazzi, Brambilla, and Nicoli]{21}
Luca Barbieri, Stefano Savazzi, Mattia Brambilla, and Monica Nicoli.
\newblock Decentralized federated learning for extended sensing in 6g connected
  vehicles.
\newblock \emph{Vehicular Communications}, 33:\penalty0 100396, 2022.

\bibitem[Wahab et~al.(2021)Wahab, Mourad, Otrok, and Taleb]{22}
Omar~Abdel Wahab, Azzam Mourad, Hadi Otrok, and Tarik Taleb.
\newblock Federated machine learning: Survey, multi-level classification,
  desirable criteria and future directions in communication and networking
  systems.
\newblock \emph{IEEE Communications Surveys \& Tutorials}, 23\penalty0
  (2):\penalty0 1342--1397, 2021.

\bibitem[Zhang et~al.(2021)Zhang, Xie, Bai, Yu, Li, and Gao]{23}
Chen Zhang, Yu~Xie, Hang Bai, Bin Yu, Weihong Li, and Yuan Gao.
\newblock A survey on federated learning.
\newblock \emph{Knowledge-Based Systems}, 216:\penalty0 106775, 2021.

\bibitem[Li et~al.(2020{\natexlab{b}})Li, Sahu, Talwalkar, and Smith]{24}
Tian Li, Anit~Kumar Sahu, Ameet Talwalkar, and Virginia Smith.
\newblock Federated learning: Challenges, methods, and future directions.
\newblock \emph{IEEE Signal Processing Magazine}, 37\penalty0 (3):\penalty0
  50--60, 2020{\natexlab{b}}.

\bibitem[Xia et~al.(2021)Xia, Ye, Tao, Wu, and Li]{25}
Qi~Xia, Winson Ye, Zeyi Tao, Jindi Wu, and Qun Li.
\newblock A survey of federated learning for edge computing: Research problems
  and solutions.
\newblock \emph{High-Confidence Computing}, 1\penalty0 (1):\penalty0 100008,
  2021.

\bibitem[Savazzi et~al.(2020)Savazzi, Nicoli, and Rampa]{26}
Stefano Savazzi, Monica Nicoli, and Vittorio Rampa.
\newblock Federated learning with cooperating devices: A consensus approach for
  massive iot networks.
\newblock \emph{IEEE Internet of Things Journal}, 7\penalty0 (5):\penalty0
  4641--4654, 2020.

\bibitem[Xiao et~al.(2017)Xiao, Rasul, and Vollgraf]{29}
Han Xiao, Kashif Rasul, and Roland Vollgraf.
\newblock Fashion-mnist: a novel image dataset for benchmarking machine
  learning algorithms.
\newblock \emph{arXiv preprint arXiv:1708.07747}, 2017.

\bibitem[Marupaka and Singh(2014)]{30}
Phani~Teja Marupaka and Rohit~Kumar Singh.
\newblock Comparison of classification results obtained by using
  cyclostationary features, mfcc, proposed algorithm and development of an
  environmental sound classification system.
\newblock In \emph{2014 International Conference on Advances in Electronics
  Computers and Communications}, pages 1--6, 2014.
\newblock \doi{10.1109/ICAECC.2014.7002428}.

\bibitem[Liu et~al.(2021{\natexlab{b}})Liu, Xu, Qin, and Tian]{27}
Xiaoyan Liu, Zhiwei Xu, Yana Qin, and Jie Tian.
\newblock A discrete-event-based simulator for distributed deep learning.
\newblock \emph{arXiv preprint arXiv:2112.00952}, 2021{\natexlab{b}}.

\end{thebibliography}

\end{document}